\documentclass[conference]{IEEEtran}
\IEEEoverridecommandlockouts
\usepackage{cite}
\usepackage{amsmath,amssymb,amsfonts}
\usepackage{algorithmic}
\usepackage{graphicx}
\usepackage{textcomp}
\usepackage{xcolor}
\usepackage[colorlinks=true, allcolors=blue]{hyperref}
\usepackage{booktabs}
\usepackage{multirow}
\usepackage{float}

\def\BibTeX{{\rm B\kern-.05em{\sc i\kern-.025em b}\kern-.08em
    T\kern-.1667em\lower.7ex\hbox{E}\kern-.125emX}}
\begin{document}

\title{Unified Diffusion-Based Rigid and Non-Rigid Editing with Text and Image Guidance\\
\thanks{\IEEEauthorrefmark{1} Denotes equal contribution.}
\thanks{\IEEEauthorrefmark{2} Denotes corresponding authors.}
\thanks{The code will be released on \href{https://github.com/Kihensarn/TI-Guided-Edit}{https://github.com/Kihensarn/TI-Guided-Edit}.}
}

\author{\IEEEauthorblockN{Jiacheng Wang\IEEEauthorrefmark{1}}
\IEEEauthorblockA{\textit{EIC, HUST} \\
jiacheng@hust.edu.cn}
\and
\IEEEauthorblockN{Ping Liu\IEEEauthorrefmark{1}}
\IEEEauthorblockA{\textit{CFAR, IHPC, A*STAR} \\
pino.pingliu@gmail.com}
\and
\IEEEauthorblockN{Wei Xu\IEEEauthorrefmark{2}}
\IEEEauthorblockA{\textit{EIC, HUST} \\
xuwei@hust.edu.cn}
}

\maketitle


\begin{figure*}[!t]
\centerline{\includegraphics[width=0.9\textwidth]{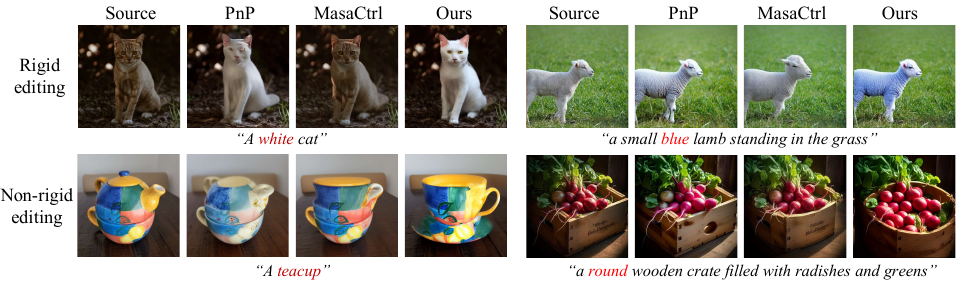}}
\caption{Illustration of limitations of prior works on rigid and non-rigid editing tasks.}
\label{fig:intro}
\end{figure*}

\begin{abstract}
Existing text-to-image editing methods tend to excel either in rigid or non-rigid editing but encounter challenges when combining both,  resulting in misaligned outputs with the provided text prompts.
In addition, integrating reference images for control remains challenging.
To address these issues, we present a versatile image editing framework capable of executing both rigid and non-rigid edits, guided by either textual prompts or reference images. 
We leverage a dual-path injection scheme to handle diverse editing scenarios and introduce an integrated self-attention mechanism for fusion of appearance and structural information. 
To mitigate potential visual artifacts, we further employ latent fusion techniques to adjust intermediate latents.
Compared to previous work, our approach represents a significant advance in achieving precise and versatile image editing.
Comprehensive experiments validate the efficacy of our method, showcasing competitive or superior results in text-based editing and appearance transfer tasks, encompassing both rigid and non-rigid settings. 
\end{abstract}
%

\section{Introduction}
\label{sec:intro}
Recent advances in text-to-image generation models\cite{ldm} have showcased impressive capabilities in generating images from natural language descriptions. 
These models enable users to create diverse objects, compositions, and scenes,  broadening the scope of image generation applications.

Recently, various methods\cite{blenddiffusion, diffedit, spatext, diffusionclip, imagic, p2p, pnp, masactrl, proxedit} have utilized text-to-image diffusion models to achieve more precise and semantically coherent editing effects. 
Some techniques like~\cite{blenddiffusion, diffedit, spatext} utilize masks automatically generated or manually provided for precise and localized image editing. 
Other approaches, such as \cite{diffusionclip, imagic}, which produce superior results, require extensive fine-tuning of the entire model. This fine-tuning process is time-consuming and can lead to overfitting.

In response to this challenge, recent works \cite{p2p, pnp, masactrl} emphasize the crucial role of attention modules in the architecture to control structural, appearance, and style information in synthesized images. 
They have designed specific attention control mechanisms that enable effective editing outcomes, eliminating the need for extensive fine-tuning.

Although these attention control methods \cite{p2p, pnp, masactrl} excel in both rigid and non-rigid editing tasks, they are not without limitations:
1) These works often focus on either rigid or non-rigid editing scenarios, making them challenging to address both simultaneously. 
As shown in Figure~\ref{fig:intro}, MasaCtrl\cite{masactrl} encounters difficulties when altering the color of a cat (rigid editing), while PnP\cite{pnp} struggles when changing the shape of a teapot (non-rigid editing). 
2) In editing scenarios that involve substantial modifications, those works may produce ambiguous results that do not align fully with the provided text prompts.
For example, as shown in Figure~\ref{fig:intro}, PnP\cite{pnp} could change the color of a cat, but there are some imperfections.
MasaCtrl \cite{masactrl} can alter the shape of a teapot as requested, but the resulting modification may not match the target text perfectly.
3) These methods primarily rely on text prompts to convey editing intentions; when incorporating a reference image as an additional control modality, they may face challenges.
For instance, it can be desirable to transfer appearance or structural information from a reference image to a source image, achieving rigid or non-rigid editing, respectively. However, these methods may struggle to transfer information from the reference image to the source image. 

To address these limitations, we present a versatile image editing framework capable of performing both rigid and non-rigid edits guided by either text or reference images.
Specifically, unlike \cite{pnp, masactrl}, our method utilizes a \textit{dual-path injection scheme} to support rigid and non-rigid editing. 
The advantage of it is to improve alignment with text prompts and to enable the injection of information from the reference image. 
To integrate appearance and structural information from different generation processes, we design a \textit{unified self-attention mechanism}, which transfers and merges information based on semantic correspondences between different images.
To alleviate potential issues such as color disparities, we introduce \textit{latent fusion} to adjust the distribution of intermediate latents.
Qualitative and quantitative evaluations demonstrate the effectiveness of our method in both text-based editing and appearance transfer tasks.

In summary, our contributions are as follows: 1) We introduce a novel dual-path injection scheme that facilitates both rigid and non-rigid editing, offering guidance options of text or reference images. 2) We propose a unified self-attention mechanism and implement different latent fusion techniques. These proposed innovations enable the integration of appearance and structural information and adjust the distribution of intermediate latent representation for enhanced visual results. 3) We conducted comprehensive qualitative and quantitative experiments, demonstrating the effectiveness of our method and showcasing comparable or superior results in both text-based editing tasks and appearance transfer tasks. 

\section{Related Work}
\textbf{Image-based Manipulation} This task aims to adapt a source image to resemble a reference image by transferring local or global semantic information.
Methods \cite{diffit, self-guidance} leverage large pre-trained diffusion models and a ``classifier" to guide noise latent distribution during diffusion, eliminating the need for additional training.
Recent work~\cite{cross-image, attn-rearrangement} uses self-attention control techniques to achieve appearance or style transfer, bridging significant domain gaps.
Our approach shares the goal of appearance transfer, but can integrate structural information into the process using a unified mechanism.

\textbf{Text-based Image Manipulation} This task aims to modify the source images according to text prompts while preserving irrelevant areas. 
The advent of diffusion models\cite{blenddiffusion, diffedit, spatext, diffusionclip, imagic} in image generation has led to various approaches that leverage these models for text-based image editing. 
To achieve impressive editing results without fine-tuning or extensive optimization,  methods\cite{p2p, pnp, masactrl}  have explored attention layers and employed attention control techniques.

Our method aligns with the paradigm of \cite{p2p, pnp, masactrl} but distinguishes itself in two key aspects: First, we employ a dual-path injection scheme that ensures the source image and target text prompt correspond to distinct generation processes. 
Secondly, we introduce a unified self-attention mechanism that \textit{concurrently} integrates appearance and structural information from multiple generation processes, aligning with the target text while preserving the source image information.

\section{Method}

\textbf{Preliminaries} To enable flexible and convenient text-guided image manipulation without fine-tuning, several methods, including Masactrl \cite{masactrl} and PnP\cite{pnp}, have adapted the self-attention mechanism within Stable Diffusion. 
Masactrl\cite{masactrl} introduces a mutual self-attention mechanism by incorporating the key features $K_{src}$ and value features $V_{src}$ from the source image into the generation process. 
This facilitates non-rigid editing, specifically altering the posture or structural information of the original image, as expressed below: 
\begin{gather}
    \mathrm{Attn}(Q,K,V)=Softmax(\frac{QK_{src}^T}{\sqrt{d}})V_{src}.
\end{gather}

PnP\cite{pnp} also suggests infusing the query features $Q_{src}$, key features $K_{src}$ and shallow spatial features $f_i$ from source image into the generated process. 
It aims for rigid editing, preserving the structural information of the source image while modifying the appearance or style details, as shown below:
\begin{gather}
    \mathrm{Attn}(Q,K,V)=Softmax(\frac{Q_{src}K_{src}^T}{\sqrt{d}})V.
\end{gather}

\begin{figure*}[!t]
\centerline{\includegraphics[width=0.75\textwidth]{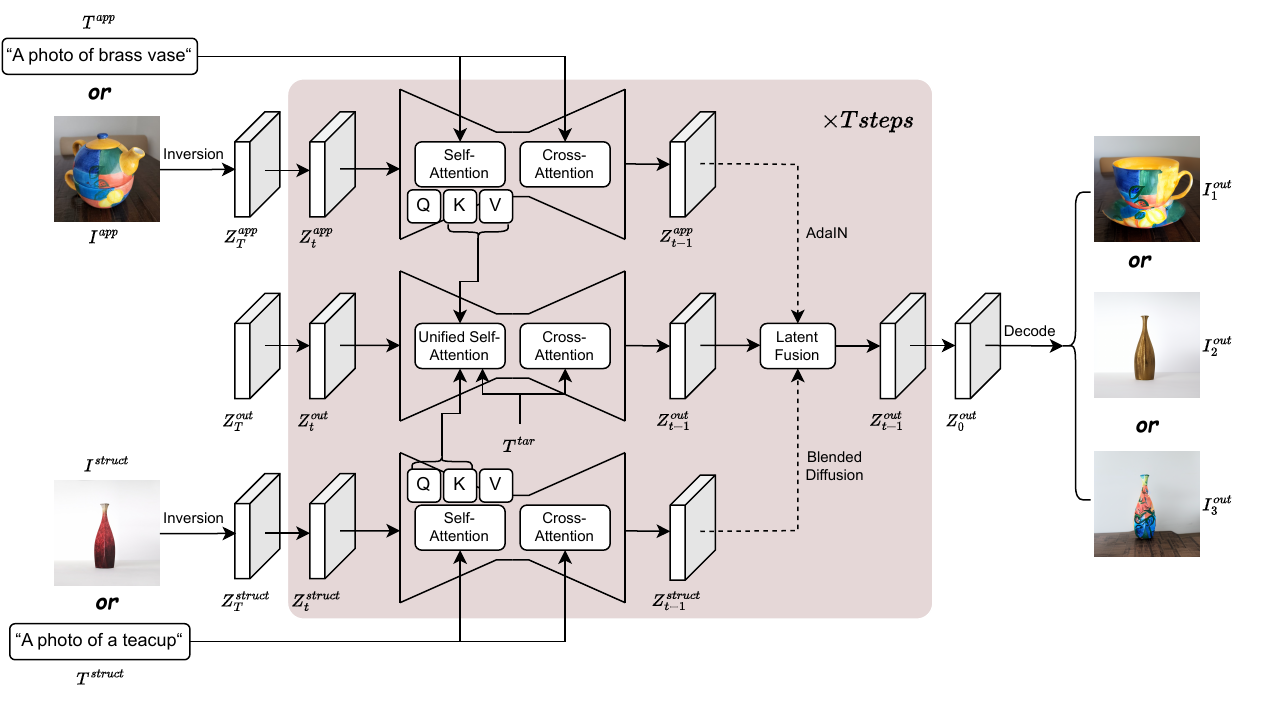}}
\caption{An Overview of Our Approach. Given the input pairs ($I^{app}$, $T^{struct}$), ($I^{struct}$, $T^{app}$), and ($I^{app}$, $I^{struct}$), our method aims to produce the desired editing result $I^{out}_1$, $I^{out}_2$, $I^{out}_2$, which corresponds to achieving text-based non-rigid edits, text-based rigid edits, and image-based edits, respectively. During the denoising step, the source image and target guidance correspond to distinct generation processes, from where information are subsequently injected into the editing process to achieve the desired manipulation.}

\label{figure_overview}
\end{figure*}

\subsection{Our Design}\label{overview}

We aim to create a versatile image editing framework for both rigid and non-rigid editing guided by text prompts or reference images. 
It allows adjustments of style, appearance, or structural elements.
We start with a source image, denoted as $I^{app}$, for non-rigid edits guided by $T^{struct}$ or precise control of structural information using $I^{struct}$. 
Conversely, starting with $I^{struct}$, we aim for rigid edits based on $T^{app}$ or utilize $I^{app}$ for control of appearance or style.

Our framework, as shown in Figure \ref{figure_overview}, employs a \textit{dual-path injection scheme} where appearance noise latent $Z_T^{app}$ and structural noise latent $Z_T^{struct}$ correspond to independent denoising processes.
These paths inject their respective information into the target noise latent $Z_t^{out}$ using a \textit{unified self-attention mechanism} throughout the denoising process.
To address potential issues such as color disparities and unwanted background changes, we introduce a \textit{latent fusion mechanism} to adjust the distribution of $Z_t^{out}$ across multiple denoising steps, ensuring alignment with editing goals.
Detailed explanations of these components follow in the subsequent sections.

\textbf{Dual-path Injection}: The most related prior works such as\cite{p2p, pnp, masactrl} employ a single-path injection scheme.
In their scheme, features from the source image are integrated into the generation process guided by the target text. 
Although this scheme facilitates text-guided edits without the need for extensive fine-tuning or optimization, it poses certain challenges.
Firstly, it retains source image details but can reduce alignment with the target text\cite{masactrl, pnp}. 
This trade-off involves preserving source information while aligning with the text\cite{masactrl, pnp}.
Second, this scheme is primarily for text guidance and is complex to extend for integrating images as precise guidance. Relying only on text may not express desired editing operations or accurately convey detailed structural, appearance, or style information.

To address the challenge of text alignment loss and significantly enhance editing flexibility, we introduce a novel dual-path injection scheme. 
In this innovative approach, the source image and the target guidance, such as a text prompt or reference image, correspond to distinct generation processes.
These two processes work harmoniously, injecting their respective information into the editing process to enable the desired manipulation. 
In other words, this introduced generation process can be harnessed to better align results with the target prompt and introduce information from the reference image, broadening our control modality to encompass both the target text and the reference image.
To simultaneously incorporate appearance and structural information from two processes within a timestep, we extend the existing self-attention mechanism, as detailed below.

Thanks to the dual-path injection scheme, we achieve various image editing configurations:
\textbf{1) Text-based non-rigid edits}: Given the source image $I^{app}$ and target prompt $T^{struct}$, we obtain $I^{out}_1$ that edits the structural information based on $T^{struct}$ while preserving the appearance information of $I^{app}$.
\textbf{2) Text-based rigid edits}: Given the source image $I^{struct}$ and target prompt $T^{app}$, we obtain $I^{out}_2$ that edits the appearance information based on $T^{app}$ while preserving the structural information of $I^{struct}$.
\textbf{3) Image-based edits}: Given the appearance image $I^{app}$ and struct image $I^{struct}$, we obtain $I^{out}_3$ that preserves the appearance information of $I^{app}$ and the structural information of $I^{struct}$.

\textbf{Unified Self-Attention Mechanism} 
As explored in \cite{p2p, pnp, masactrl}, the self-attention layers in the denoising U-Net contain a substantial amount of structure, appearance, and style information for generating images. 
By modifying queries, keys, or values in these layers, we can selectively influence specific aspects of the generated images, such as style, texture, or layouts, enabling both rigid and non-rigid editing.
Specifically, queries and keys yield an attention map, representing the similarity and semantic correspondence between each query and the rest of the image. 
Values represent the content features in the synthesis process, which are aggregated based on the semantic correspondence in the attention map. 

Building upon these concepts, we propose a unified self-attention mechanism designed to consolidate appearance and structural information sourced from different images. 
Specifically, for the appearance image $I_{app}$, we define its query features, key features, and value features as $Q_{app}$, $K_{app}$, and $V_{app}$; for the struct image $I_{struct}$, we define them as $Q_{struct}$, $K_{struct}$, and $V_{struct}$; for the editing image, we define them as $Q_{out}$, $K_{out}$, and $V_{out}$. For simplicity, we omit the timestep ($t$) and layer ($l$) notation in the discussion here.
Concretely, our unified self-attention mechanism is computed through the following process.
First, we compute the structural attention map, representing the semantic correspondences within $I_{struct}$, which can be expressed as:
\begin{gather}
Attn_{struct}=Softmax(\frac{Q_{struct}K_{struct}^T}{\sqrt{d}}).
\end{gather}
Second, based on $Q_{out}$ and $K_{app}$, we can obtain a similarity matrix $Sim_{out}=Q_{out}K_{app}^T$, which reflects the semantic similarity between $I_{out}$ and $I_{app}$.

Thirdly, we reweight $Sim_{out}$ by $Attn_{struct}$, obtaining a unified attention map, which retains the structural information of $I_{struct}$ while incorporating semantic correspondences with $I_{app}$. This can be expressed as:
\begin{gather}
Attn_{uni}=Softmax(\frac{Attn_{struct}Sim_{out}}{\sqrt{d}}).
\end{gather}
Finally, by multiplying the unified attention map with the value features from the appearance image, we can precisely inject the appearance information from $I_{app}$ into $I_{out}$, under the structural guidance of $I_{struct}$.

However, we observed that the straightforward application of the unified self-attention mechanism occasionally led to structural mismatches or undesired artifacts in specific scenarios. 
To solve this issue, we introduce the contrast\cite{cross-image} and rearrange\cite{attn-rearrangement} operations into our unified self-attention, enhancing semantic alignments and improving generative quality.
The contrast operation is defined as: $ \mathcal{F}_c(\boldsymbol{X})=(\boldsymbol{X}-\mu(\boldsymbol{X}))\beta+\mu(\boldsymbol{X})$,
where $\mu$ represents the mean operation, and $\beta$ denotes the contrast factor.
The rearrange operation is formulated as: $    \mathcal{F}_r(\boldsymbol{X}, \boldsymbol{Y})=Softmax(\begin{bmatrix}\frac{\boldsymbol{X}}{\sqrt d}+C,\frac{\boldsymbol{Y}}{\sqrt d}\end{bmatrix})$,
where $C=\ln\frac{\sum_j\exp\left([\boldsymbol{Y}]_{\cdot,j}\right)}{\sum_j\exp\left([\boldsymbol{X}]_{\cdot,j}\right)}$.

The final unified self-attention with attention contrast and rearrangement is illustrated as:
\begin{gather}                            Attn(Q,K,V)=\mathcal{F}_c(Attn'_{uni})\begin{bmatrix}V_{app} \\ V_{out}\end{bmatrix},
\end{gather}
where $Attn'_{uni}=\mathcal{F}_r(Attn'_{struct}(Q_{out}K_{app}^T), Q_{out}K_{out}^T)$, $Attn'_{struct}=\mathcal{F}_c(Softmax(\frac{Q_{struct}K_{struct}^T}{\sqrt{d}}))$.

To inject structural and appearance information from different generation processes into the editing process, we apply this mechanism in multiple time steps and layers during the iterative denoising steps.

\textbf{Latent Fusion} In image-based editing scenarios, we observed that a notable dissimilarity between the appearance image and the structural image could lead to pronounced color shifts, blurring, and haziness. 
We speculate that this issue arises from the challenge of identifying precise semantic correspondences when computing the self-attention map across two distinct images.
To address this, we adopt the AdaIN\cite{adain} technique following the approach in \cite{cross-image}.
AdaIN is used to match the feature distributions of $z_{app}$ and $z_{out}$, alleviating color shift problems.
This operation can be expressed as: 
\begin{equation}
\begin{aligned}
    z_{out}&=(1-m_{struct})*z_{out} \\ 
    &+m_{struct}*\text{AdaIN}(z_{out},z_{app}*m_{app}),
\end{aligned}
\end{equation}
where $m_{app}$ and $m_{struct}$ are the foreground masks of the appearance image and struct image, respectively. 

In the context of rigid editing, we also observed substantial background variations in the generated results, which are not in line with our expectations. 
To address this problem, we employ the blended diffusion strategy\cite{blenddiffusion}, fusing $z_{out}$ and $z_{struct}$ with the help of $m_{struct}$. 
\begin{equation}
\begin{aligned}
    z_{out}&=m_{struct}*z_{out}+(1-m_{struct})*z_{struct},
\end{aligned}
\end{equation}
where $m_{struct}$ is the foreground mask of the struct image. These masks are automatically computed from cross-attention maps during the inversion process\cite{p2p}.

Leveraging all the mechanisms mentioned above, our method empowers rigid and non-rigid editing guided by text prompts or reference images in nearly 50 sampling steps, all without requiring the burden of fine-tuning or optimization.

\section{Experiments}

\textbf{Implement Details} We employ the pre-trained text-to-image Stable Diffusion v1.5\cite{ldm} as our generative model and employ Negative-Prompt Inversion\cite{negative} to invert real images in $50$ steps. 
During the editing process, we adopted the DDIM sampling strategy\cite{ddim} over $50$ steps. 
Appearance information is injected between timesteps $4$ and $40$, while structural information is injected between timesteps $0$ and $25$~\footnote{We have the option to extend up to $35$ steps for enhanced preservation of structural information}.
For the image-based editing setting, we apply AdaIN from $10$ to $30$ timesteps. 
For the text-based rigid editing setting, we employ the blended diffusion mechanism from timesteps $0$ to $40$.
\begin{figure}[!t]
\centerline{\includegraphics[width=0.45\textwidth]{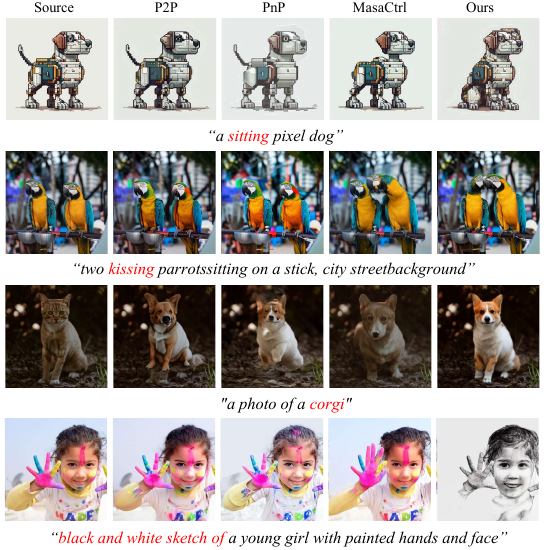}}
\caption{Qualitative Comparisons with Text-Based Editing Methods: P2P\cite{p2p}, PnP\cite{pnp}, MasaCtrl\cite{masactrl}. Our approach successfully accomplishes both rigid and non-rigid editing, demonstrating improved alignment with the target prompt.}
\label{fig:text_comparision}
\end{figure}

\textbf{Comparisons with Text-based Editing Methods} We compare three text-based image editing methods that utilize self-attention control: P2P\cite{p2p}, PnP\cite{pnp}, and MasaCtrl\cite{masactrl}. 
We conducted quantitative and qualitative experiments on PIE-Bench\cite{direct_inversion}, which covers nine different editing scenarios, including rigid and non-rigid, local and global editing tasks.
The Benchmark offers a comprehensive evaluation framework that considers factors such as Structure Preservation, Background Preservation, and CLIP Similarity Score of the whole image and regions in the editing mask.

Qualitative results are shown in Figure \ref{fig:text_comparision}. 
For the non-rigid editing tasks depicted in the first two rows, both P2P and PnP face challenges in editing the pose or shape according to the target prompt, indicating limitations in handling such editing scenarios. 
Although MasaCtrl is designed for non-rigid editing tasks, it still encounters difficulties in achieving desired changes and is outperformed by our solution.
Regarding the rigid editing tasks illustrated in the last two rows, MasaCtrl appears to perform unrelated editing operations, suggesting its unsuitability for rigid editing scenarios.
In contrast, although P2P and PnP manage to achieve the required editing effects, their results may not fully align with the provided prompt. 
By utilizing our dual-path injection scheme and unified self-attention mechanism, we not only address both rigid and non-rigid editing scenarios but also minimize instances of ambiguous editing outcomes.

\begin{table*}[!t]
\footnotesize
\caption{Quantitative Comparisons to Text-based Editing Methods. In each evaluation metric, we  highlighted the top performing method in bold and the second best performing method with an \underline{underscore}. We average the results from both our non-rigid and rigid editing settings. Our approach consistently achieves comparable results across all evaluation metrics.}
\label{table:quantitative_comparison_text}
\centering
\setlength{\tabcolsep}{2mm}{
\begin{tabular}{c|c|cccc|cc}
\toprule
\multicolumn{1}{c}{\multirow{2}{*}{}} & \multicolumn{1}{c}{\textbf{Structure}}     & \multicolumn{4}{c}{\textbf{Background Preservation}}   & \multicolumn{2}{c}{\textbf{CLIP Similarity}} \\
\midrule
\textbf{Method} & $\textbf{Distance}_{\times10^3}\downarrow$ & \textbf{PSNR}$\uparrow$ & $\textbf{LPIPS}_{\times10^3}\downarrow$ & $\textbf{MSE}_{\times10^4}\downarrow$ & $\textbf{SSIM}_{\times10^2}\uparrow$ & \textbf{Whole}$\uparrow$  & \textbf{Edited}$\uparrow$ \\
\midrule
P2P\cite{p2p}      & \textbf{11.65}  & \textbf{27.22} & \textbf{54.55}  & \textbf{32.86}  & \textbf{84.76}  & 25.02 & 22.10  \\
MasaCtrl\cite{masactrl} & 24.70   & 22.64 & 87.94  & 81.09  & 81.33  & 24.38 & 21.35 \\
PnP\cite{pnp}     & 24.29  & 22.46 & 106.06 & 80.45  & 79.68  & \textbf{25.41} & \textbf{22.62} \\
Ours &\underline{22.13}	&\underline{24.74}	&\underline{84.45}	&\underline{58.45}	&\underline{81.62}	&\underline{25.15}	&\underline{22.12} \\
\bottomrule
\end{tabular}}
\end{table*}

\begin{figure}[!t]
\centerline{\includegraphics[width=0.45\textwidth]{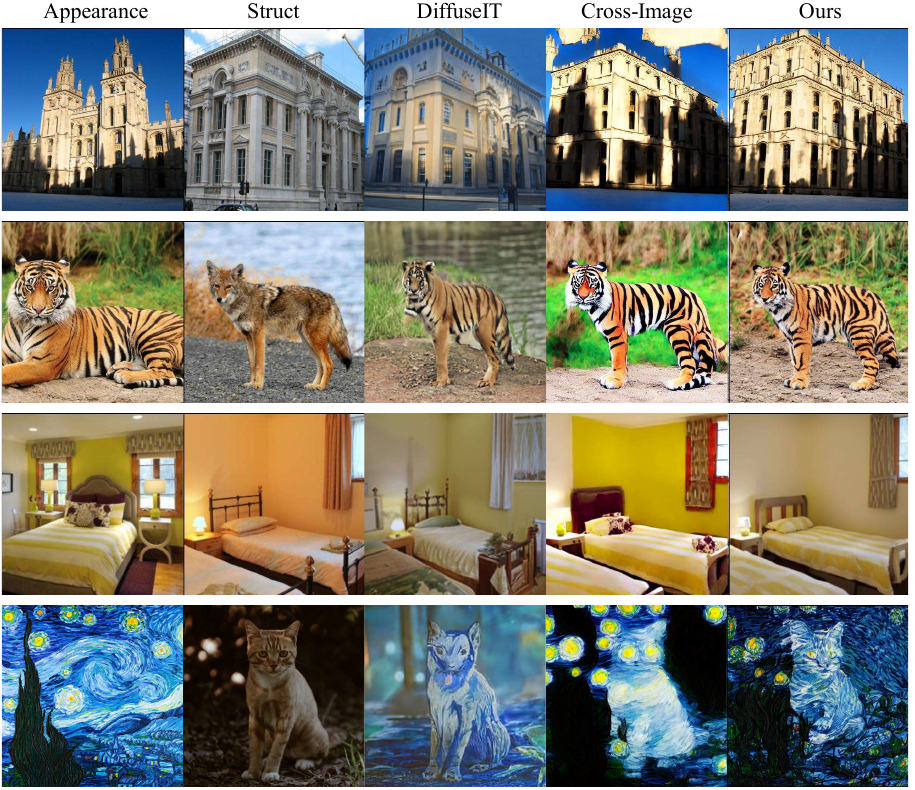}}
\caption{Qualitative Comparisons to Appearance Transfer Methods. Our method can effectively integrate appearance and structural information from different images.}
\label{fig:image_comparision}
\end{figure}

\begin{figure}[!t]
\centerline{\includegraphics[width=0.5\textwidth]{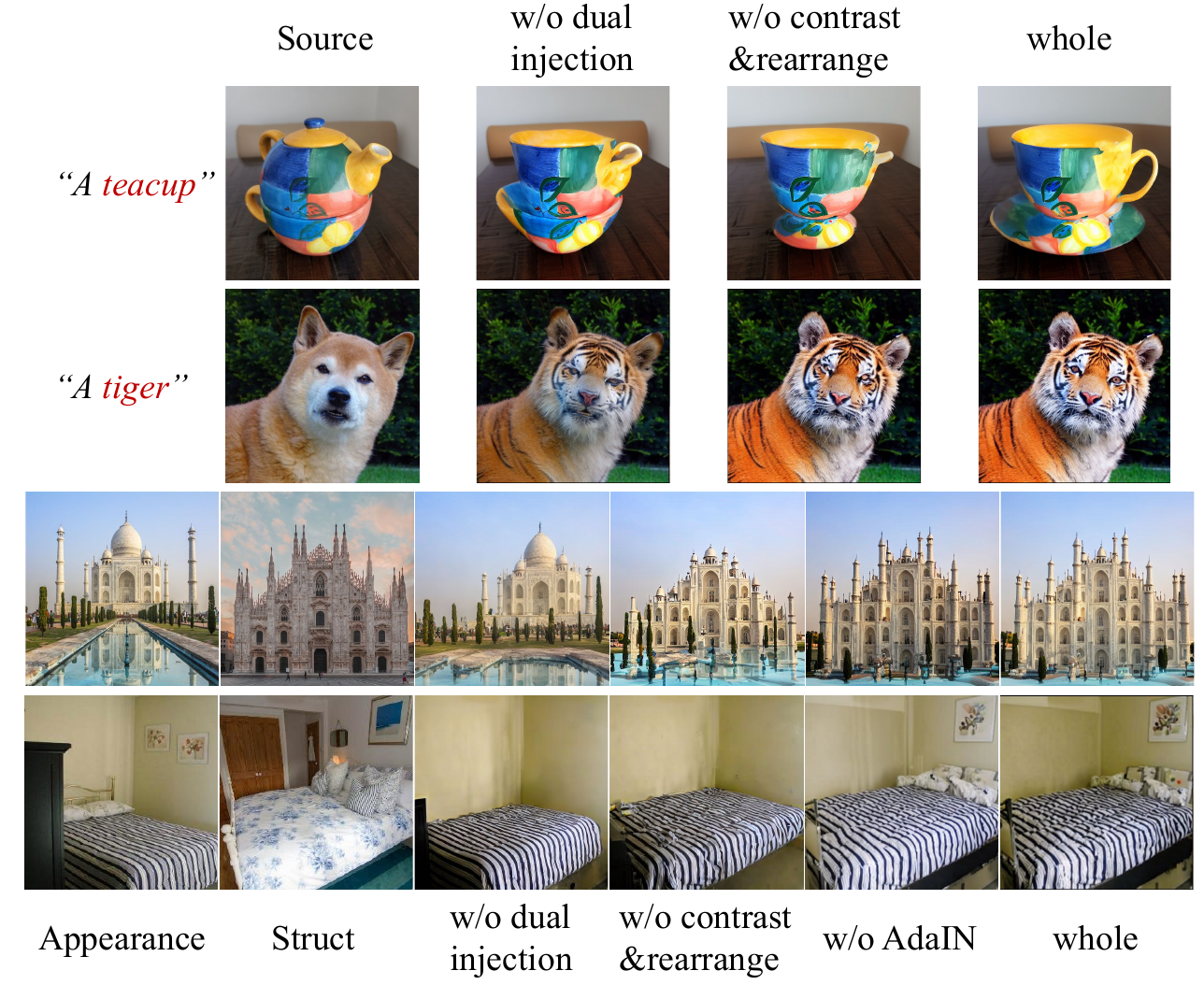}}
\caption{Ablation Study. Each column represents the effect of removing a specific component.}
\label{fig:ablation}
\end{figure}

In Table \ref{table:quantitative_comparison_text}, we present the quantitative results, averaging the results from our non-rigid and rigid edit settings with dual-path injection every $4$ timesteps.
P2P exhibits remarkable structural and background preservation capabilities. 
However, it demands a high level of alignment between the source and target prompts, which may not always be achievable and can be somewhat limiting for practical use.
PnP achieves the highest CLIP Similarity, showcasing its effectiveness in certain aspects. However, it comes with a compromise in background preservation.
MasaCtrl obtains slightly lower evaluation results, but it excels in handling non-rigid edits, which may pose challenges for P2P and PnP.
Our method, although not achieving the highest evaluation scores, still delivers competitive performance.
It demonstrates noteworthy capabilities in both rigid and non-rigid editing scenarios, providing a versatile and flexible tool for image editing tasks.
While our scores may not be the highest, our method's versatility and proficiency in various editing scenarios make it a valuable addition to the field.

\textbf{Comparisons with Appearance Transfer Methods} We compare our method with state-of-the-art methods based on large-scale pre-trained diffusion models: DiffuseIT\cite{diffit} and Cross-Image\cite{cross-image}. 
As shown in Figure \ref{fig:image_comparision}, DiffuseIT effectively preserves structural information but may incur losses in appearance information in certain cases, such as the church image in the first row. 
Cross-Image demonstrates good preservation of appearance information, but often exhibits noticeable color shifts, with slightly lower structural preservation capabilities. 
Our method, powered by our unified self-attention mechanism, uniquely integrates structural information into the image generation process. 
This integration significantly enhances the preservation of both the appearance and structural details.
Moreover, as shown in the last row, we explore our ability of style transfer. 
These results demonstrate the competence of our method in retaining stylistic and structural characteristics and achieving better results.

\begin{table*}[!t]
\footnotesize
{\caption{Quantitative Comparisons to Appearance Transfer Methods. We achieve superior results on both metrics, demonstrating robust preservation capabilities for both structural and appearance information.}}
\label{table:quantitative_comparison_image}
\centering
\begin{tabular}{c|ccc|ccc}
\toprule
\multicolumn{1}{c}{\multirow{2}{*}{}} & \multicolumn{3}{c}{$\textbf{Structure Distance}_{\times10^2}\downarrow$} & \multicolumn{3}{c}{$\textbf{Appearance Distance}_{\times10^2}\downarrow$} \\
\midrule
Domain & DiffuseIT\cite{diffit} & Cross-Image\cite{cross-image} & Ours & DiffuseIT\cite{diffit} & Cross-Image\cite{cross-image} & Ours \\
\midrule
Animal Faces & 8.37 & 12.80 & 10.01 & 2.71 & 4.59 & 3.13 \\
Animals & 8.29 & 10.46 & 9.21 & 5.64 & 3.96 & 3.18 \\
Bedroom & 9.41 & 11.99 & 7.68 & 4.17 & 5.29 & 2.07 \\
Cars & 8.65 & 10.17 & 6.95 & 5.54 & 4.70 & 2.08 \\
Churches & 8.02 & 8.59 & 5.93 & 4.63 & 4.30 & 4.58 \\
\midrule
Average & 8.55  & 10.80 & \textbf{7.96} & 4.54 & 4.57 & \textbf{3.01} \\
\bottomrule
\end{tabular}
\end{table*}
We quantitatively evaluate the performance on: 
1) Structural Preservation: To gauge the preservation of structural information, we employ a structural distance metric following the methodology described in \cite{direct_inversion}.
2) Appearance Preservation: To assess the preservation of appearance information, we utilized a pre-trained VGG19 network that can extract intermediate features from both the generated results and the source appearance images.
We measure preservation quality by calculating the L2 distance between their respective Gram matrices.
As there is no universal evaluation dataset, following \cite{cross-image}, we conduct evaluation on five domains (animal faces, animals, bedrooms, cars, and churches), where we select $20$ data pairs from each.
As shown in Table \ref{table:quantitative_comparison_image}, our method demonstrates superior performance.

\textbf{Ablation Study} Our ablation study further validates the effectiveness of our method by examining key components. Specifically, as shown in Figure \ref{fig:ablation}, we investigate:
1) {Dual-Path Injection Scheme}: Its removal presents challenges in aligning with text prompts and preserving structural information.
This underscores its role in enhancing text-image alignment and integrating various information sources.
2) {Contrast and Rearrange Operations}: When these operations are omitted from the unified self-attention mechanism, artifacts emerge (\textit{e.g.}, bedsheet patterns) alongside structural information loss (e.g., teacup and church shapes).
This confirms its role in improving semantic correspondence and precise fusion of appearance and structural data.
3) {AdaIN Normalization}: Its absence results in minor color distortions, particularly in certain scenarios (\textit{e.g.}, church image), highlighting its effectiveness in mitigating color distortion issues.

\section{Conclusion}

We introduce a versatile image editing framework to handle both rigid and non-rigid edits, guided by text or reference images. 
We propose a dual-path scheme, unified self-attention, and latent fusion to fuse the appearance and structure of different resources. 
Extensive evaluations demonstrate the competitive or superior performance of our method.

\bibliographystyle{IEEEtran}
\bibliography{ref}

\begin{thebibliography}{10}
\providecommand{\url}[1]{#1}
\csname url@samestyle\endcsname
\providecommand{\newblock}{\relax}
\providecommand{\bibinfo}[2]{#2}
\providecommand{\BIBentrySTDinterwordspacing}{\spaceskip=0pt\relax}
\providecommand{\BIBentryALTinterwordstretchfactor}{4}
\providecommand{\BIBentryALTinterwordspacing}{\spaceskip=\fontdimen2\font plus
\BIBentryALTinterwordstretchfactor\fontdimen3\font minus \fontdimen4\font\relax}
\providecommand{\BIBforeignlanguage}[2]{{%
\expandafter\ifx\csname l@#1\endcsname\relax
\typeout{** WARNING: IEEEtran.bst: No hyphenation pattern has been}%
\typeout{** loaded for the language `#1'. Using the pattern for}%
\typeout{** the default language instead.}%
\else
\language=\csname l@#1\endcsname
\fi
#2}}
\providecommand{\BIBdecl}{\relax}
\BIBdecl

\bibitem{ldm}
R.~Rombach, A.~Blattmann, D.~Lorenz, P.~Esser, and B.~Ommer, ``High-resolution image synthesis with latent diffusion models,'' in \emph{CVPR}, 2022.

\bibitem{blenddiffusion}
O.~Avrahami, D.~Lischinski, and O.~Fried, ``Blended diffusion for text-driven editing of natural images,'' in \emph{CVPR}, 2022.

\bibitem{diffedit}
G.~Couairon, J.~Verbeek, H.~Schwenk, and M.~Cord, ``Diffedit: Diffusion-based semantic image editing with mask guidance,'' in \emph{ICLR}, 2023.

\bibitem{spatext}
O.~Avrahami, T.~Hayes, O.~Gafni, S.~Gupta, Y.~Taigman, D.~Parikh, D.~Lischinski, O.~Fried, and X.~Yin, ``Spatext: Spatio-textual representation for controllable image generation,'' in \emph{CVPR}, 2023.

\bibitem{diffusionclip}
G.~Kim, T.~Kwon, and J.~C. Ye, ``Diffusionclip: Text-guided diffusion models for robust image manipulation,'' in \emph{CVPR}, 2022.

\bibitem{imagic}
B.~Kawar, S.~Zada, O.~Lang, O.~Tov, H.~Chang, T.~Dekel, I.~Mosseri, and M.~Irani, ``Imagic: Text-based real image editing with diffusion models,'' in \emph{CVPR}, 2023.

\bibitem{p2p}
A.~Hertz, R.~Mokady, J.~Tenenbaum, K.~Aberman, Y.~Pritch, and D.~Cohen{-}Or, ``Prompt-to-prompt image editing with cross-attention control,'' in \emph{ICLR}, 2023.

\bibitem{pnp}
N.~Tumanyan, M.~Geyer, S.~Bagon, and T.~Dekel, ``Plug-and-play diffusion features for text-driven image-to-image translation,'' in \emph{CVPR}, 2023.

\bibitem{masactrl}
M.~Cao, X.~Wang, Z.~Qi, Y.~Shan, X.~Qie, and Y.~Zheng, ``Masactrl: Tuning-free mutual self-attention control for consistent image synthesis and editing,'' in \emph{ICCV}, 2023.

\bibitem{proxedit}
L.~Han, S.~Wen, Q.~Chen, Z.~Zhang, K.~Song, M.~Ren, R.~Gao, A.~Stathopoulos, X.~He, Y.~Chen \emph{et~al.}, ``Proxedit: Improving tuning-free real image editing with proximal guidance,'' in \emph{WACV}, 2024.

\bibitem{diffit}
G.~Kwon and J.~C. Ye, ``Diffusion-based image translation using disentangled style and content representation,'' in \emph{ICLR}, 2023.

\bibitem{self-guidance}
D.~Epstein, A.~Jabri, B.~Poole, A.~A. Efros, and A.~Holynski, ``Diffusion self-guidance for controllable image generation,'' in \emph{NeurIPS}, 2023.

\bibitem{cross-image}
Y.~Alaluf, D.~Garibi, O.~Patashnik, H.~Averbuch-Elor, and D.~Cohen-Or, ``Cross-image attention for zero-shot appearance transfer,'' \emph{arXiv}, 2023.

\bibitem{attn-rearrangement}
Y.~Deng, X.~He, F.~Tang, and W.~Dong, ``Zero-shot style transfer via attention rearrangement,'' \emph{arXiv}, 2023.

\bibitem{adain}
X.~Huang and S.~Belongie, ``Arbitrary style transfer in real-time with adaptive instance normalization,'' in \emph{ICCV}, 2017.

\bibitem{negative}
D.~Miyake, A.~Iohara, Y.~Saito, and T.~Tanaka, ``Negative-prompt inversion: Fast image inversion for editing with text-guided diffusion models,'' \emph{arXiv}, 2023.

\bibitem{ddim}
J.~Song, C.~Meng, and S.~Ermon, ``Denoising diffusion implicit models,'' in \emph{ICLR}, 2021.

\bibitem{direct_inversion}
X.~Ju, A.~Zeng, Y.~Bian, S.~Liu, and Q.~Xu, ``Direct inversion: Boosting diffusion-based editing with 3 lines of code,'' \emph{arXiv}, 2023.

\end{thebibliography}


\section*{Supplemental Material}

\begin{figure*}[h]
    \centering
    \includegraphics{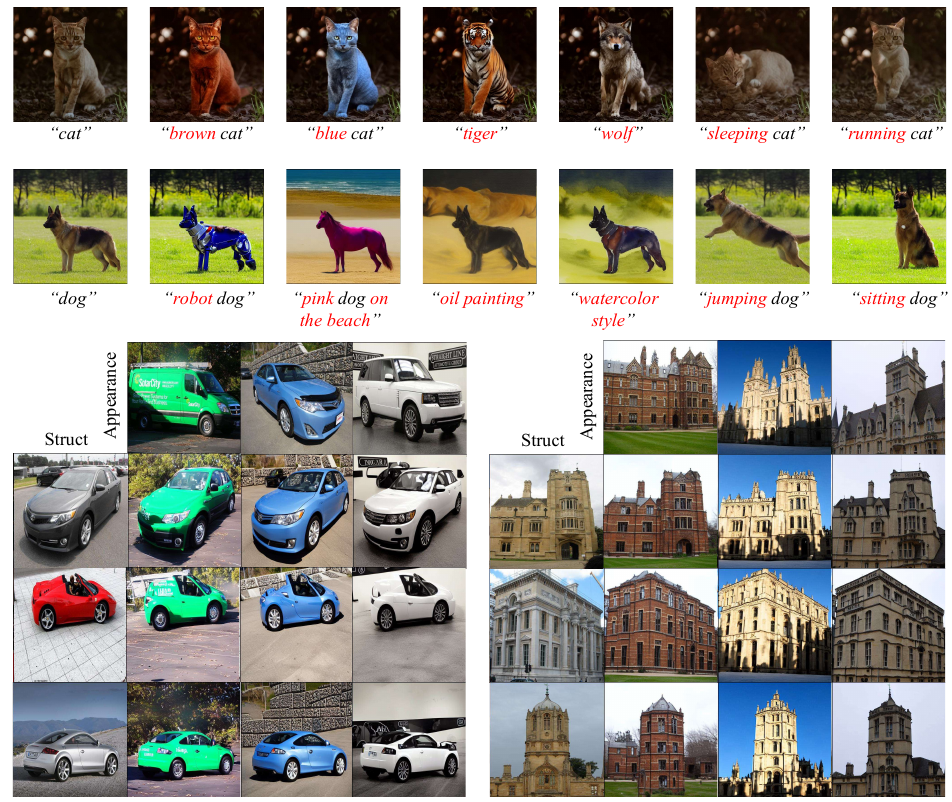}
    \caption{Our approach demonstrates strong performance in both rigid and non-rigid editing scenarios, guided by text prompts or reference images.}
    \label{fig:supplement_overview}
\end{figure*}

We provide more experimental results to demonstrate the overall effectiveness of our method. 
As illustrated in Figure \ref{fig:supplement_overview}, our approach showcases its ability to control object pose, color, species, and style based on text prompts. 
Additionally, our approach is capable of manipulating the appearance or structural information of a source image based on a reference image. 

\begin{figure*}[h]
    \centering
    \includegraphics{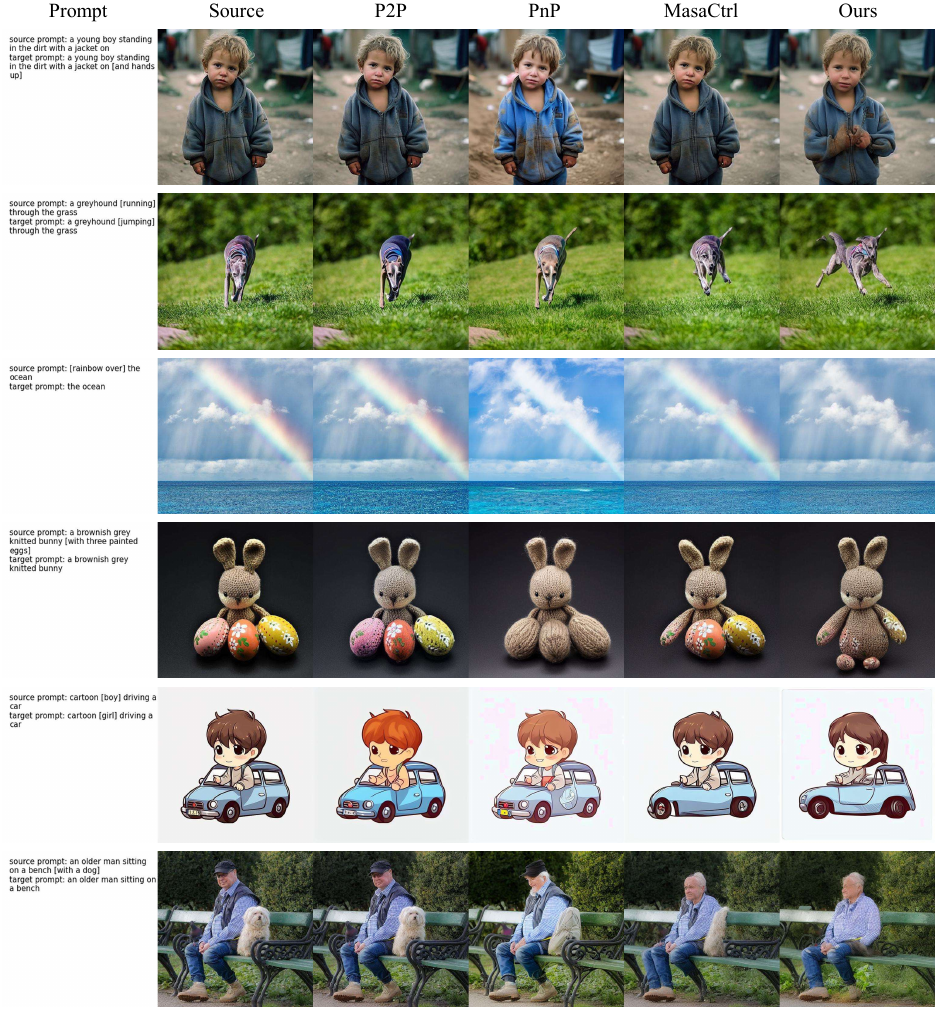}
    \caption{Additional qualitative comparison on text-based non-rigid editing.}
    \label{fig:supplement_text_comparison_1}
\end{figure*}

\begin{figure*}[h]
    \centering
    \includegraphics{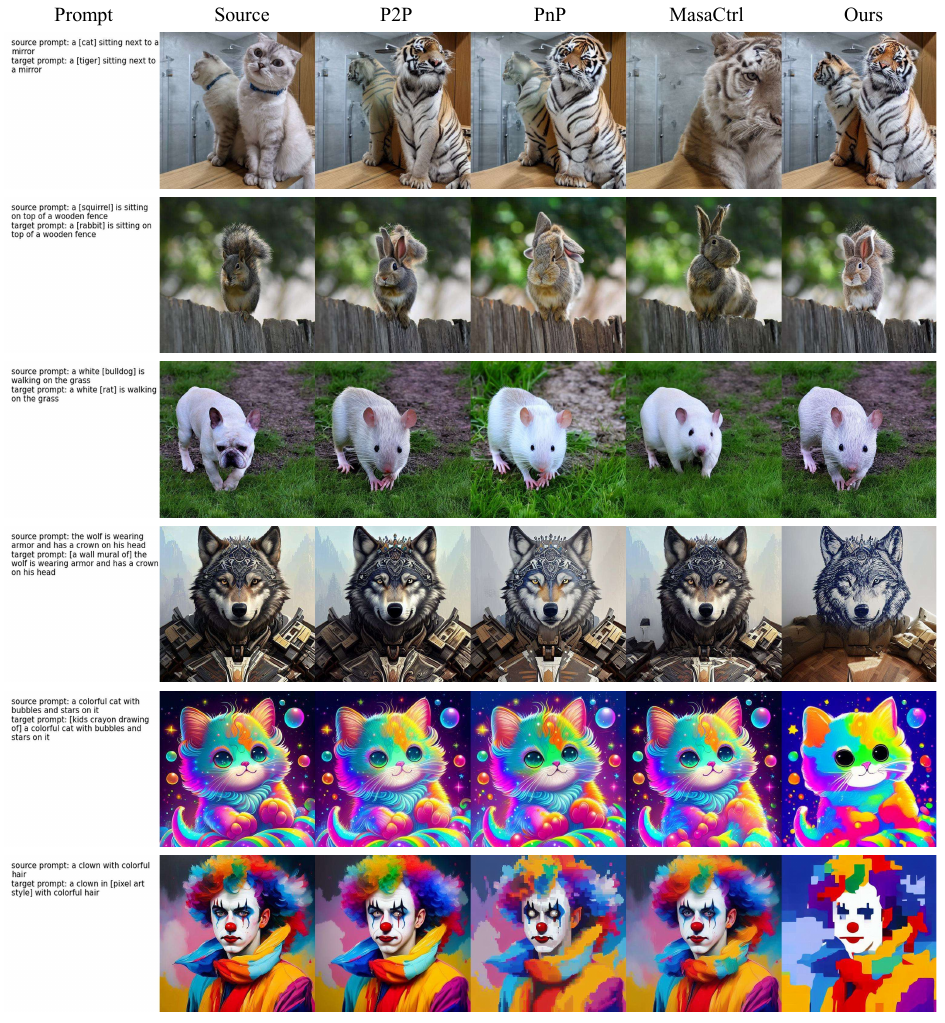}
    \caption{Additional qualitative comparison on text-based rigid editing.}
    \label{fig:supplement_text_comparison_2}
\end{figure*}

\begin{figure*}[h]
    \centering
    \includegraphics{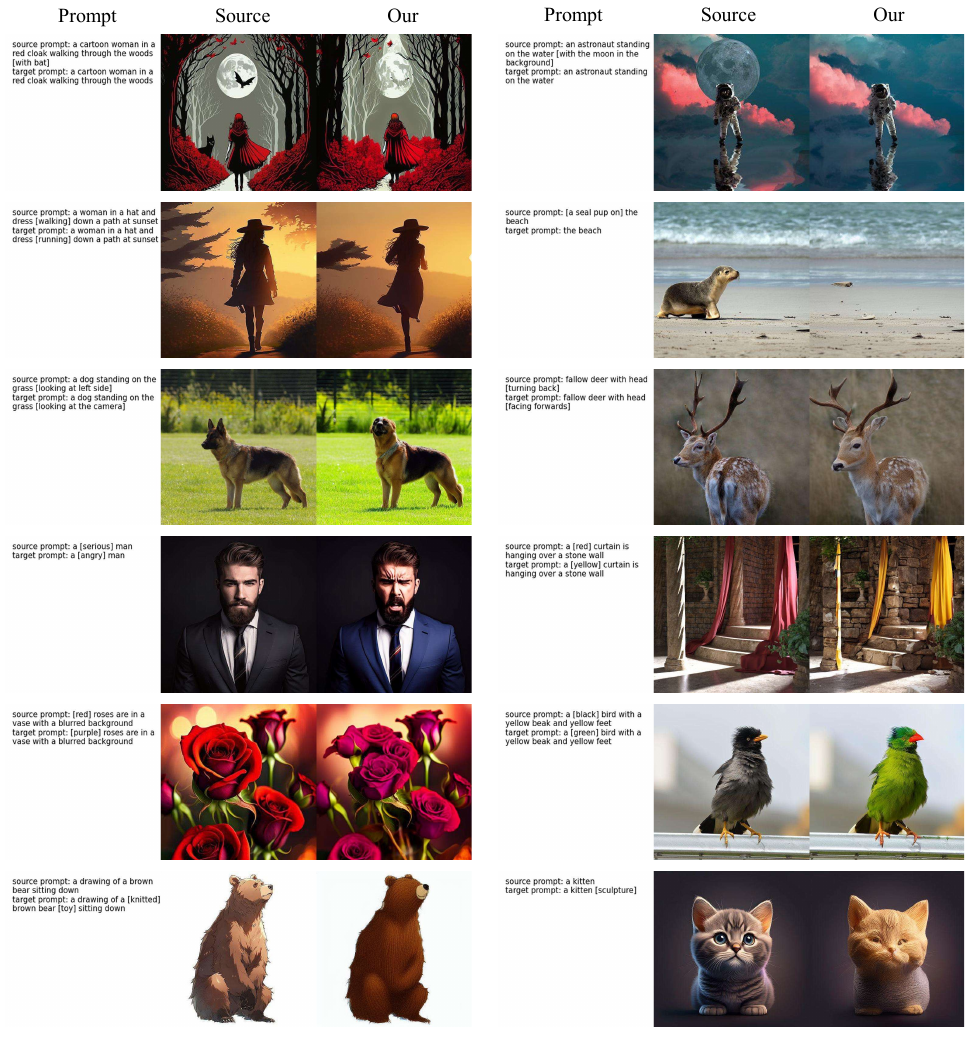}
    \caption{Additional visual results of our approach in various editing scenarios}
    \label{fig:supplement_text_vis}
\end{figure*}

In the context of text-based editing tasks, we present more qualitative results in Figure \ref{fig:supplement_text_comparison_1} and Figure \ref{fig:supplement_text_comparison_2}. 
As shown in the figures, our method adeptly manages both rigid and non-rigid edits, consistently achieving comparable or superior results across various editing scenarios.
Further visual results, as shown in Figure \ref{fig:supplement_text_vis}, highlight the effectiveness of our approach in handling complex editing scenarios.

\begin{figure*}[h]
    \centering
    \includegraphics[width=0.9\textwidth]{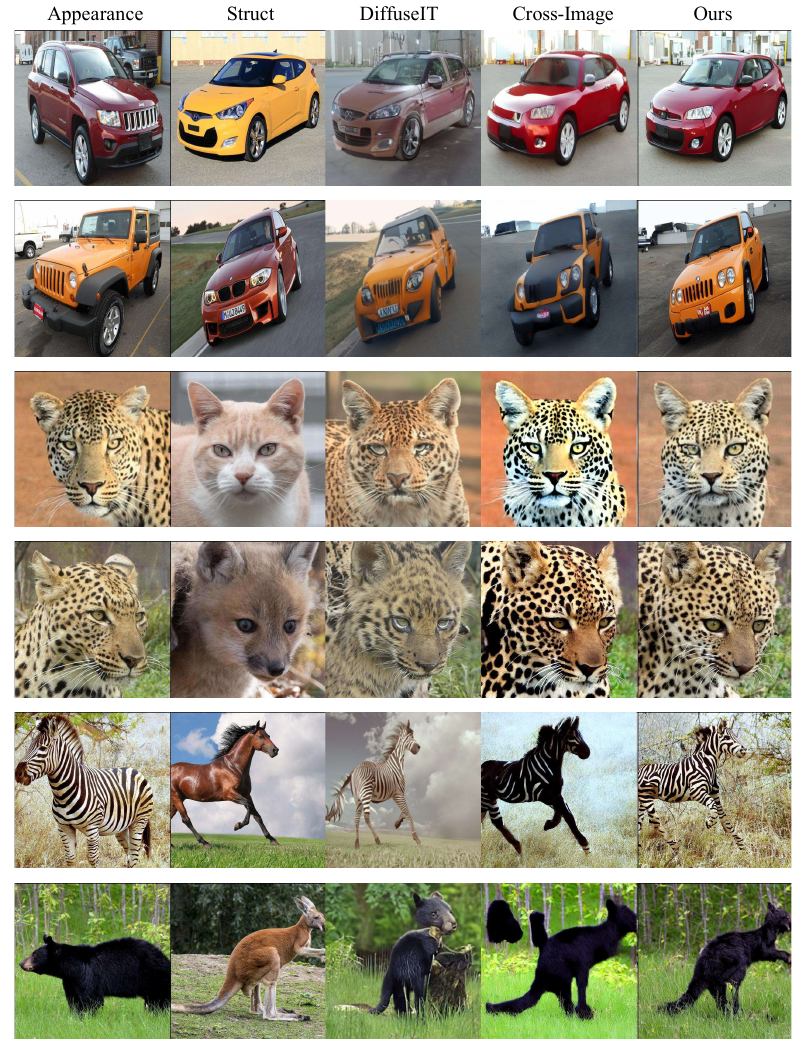}
    \caption{Additional qualitative comparison on appearance transfer task.}
    \label{fig:supplement_img_comparison_1}
\end{figure*}

\begin{figure*}[h]
    \centering
    \includegraphics[width=0.9\textwidth]{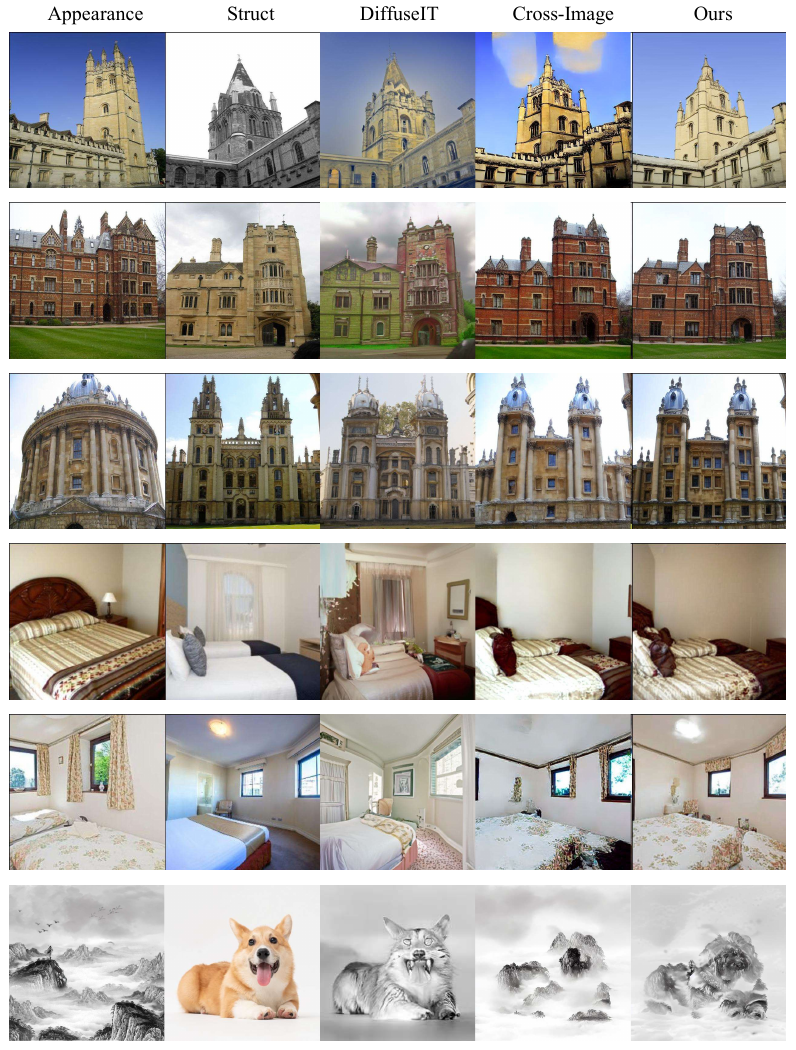}
    \caption{Additional qualitative comparison on appearance transfer task.}
    \label{fig:supplement_img_comparison_2}
\end{figure*}

\begin{figure*}[h]
    \centering
    \includegraphics[width=0.9\textwidth]{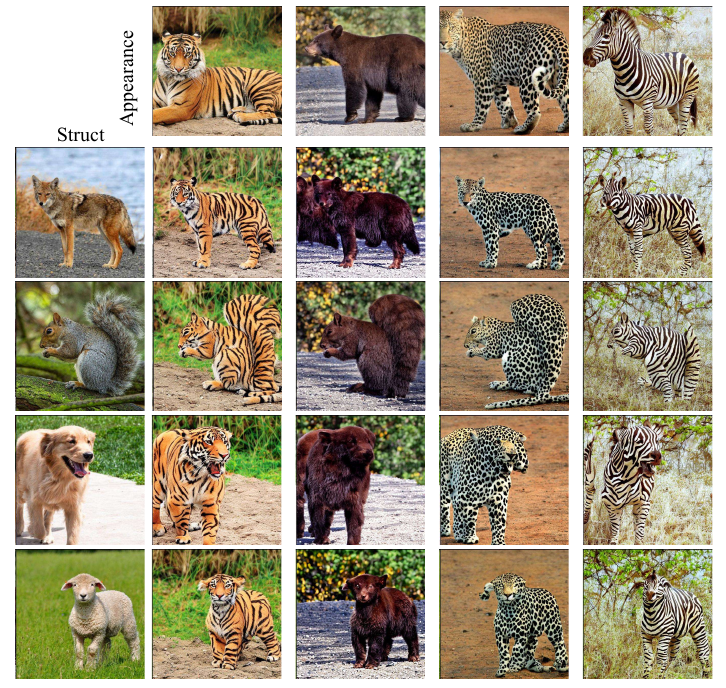}
    \caption{Additional visual results of our approach in animal domain.}
    \label{fig:supplement_img_vis_1}
\end{figure*}

\begin{figure*}[h]
    \centering
    \includegraphics[width=0.9\textwidth]{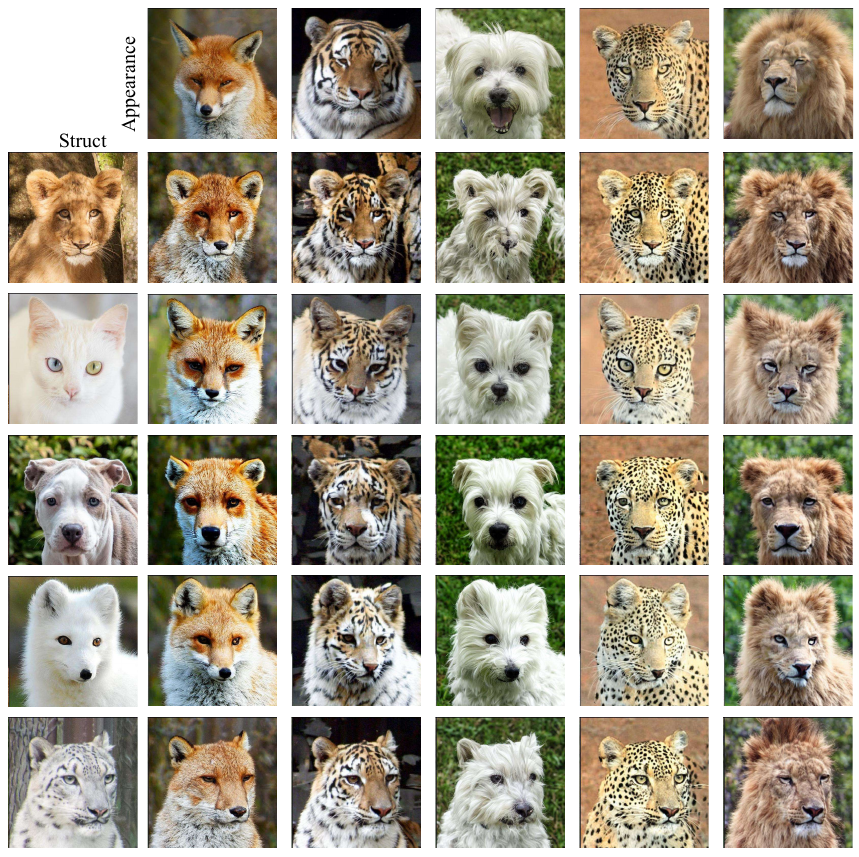}
    \caption{Additional visual results of our approach in animal face domain.}
    \label{fig:supplement_img_vis_2}
\end{figure*}

For appearance transfer tasks, additional qualitative experimental results are provided in Figure \ref{fig:supplement_img_comparison_1} and Figure \ref{fig:supplement_img_comparison_2}.
The results demonstrate the consistent outperformance of our method across multiple domains.
More visual results, particularly focusing on animals and animal faces, as illustrated in Figure \ref{fig:supplement_img_vis_1} and Figure \ref{fig:supplement_img_vis_2}, showcase the better performance of our method even in the presence of significant pose variations among animals.

\end{document}